# Autonomous object harvesting using synchronized optoelectronic microrobots*

Christopher Bendkowski, Laurent Mennillo, Tao Xu, Mohamed Elsayed, Filip Stojic, Harrison Edwards, Shuailong Zhang, Cindi Morshead, Vijay Pawar, Aaron R. Wheeler, Danail Stoyanov, Michael Shaw

*Abstract*—Optoelectronic tweezer-driven microrobots (OETdMs) are a versatile micromanipulation technology based on the use of light induced dielectrophoresis to move small dielectric structures (microrobots) across a photoconductive substrate. The microrobots in turn can be used to exert forces on secondary objects and carry out a wide range of micromanipulation operations, including collecting, transporting and depositing microscopic cargos. In contrast to alternative (direct) micromanipulation techniques, OETdMs are relatively gentle, making them particularly well suited to interacting with sensitive objects such as biological cells. However, at present such systems are used exclusively under manual control by a human operator. This limits the capacity for simultaneous control of multiple microrobots, reducing both experimental throughput and the possibility of cooperative multi-robot operations. In this article, we describe an approach to automated targeting and path planning to enable open-loop control of multiple microrobots. We demonstrate the performance of the method in practice, using microrobots to simultaneously collect, transport and deposit silica microspheres. Using computational simulations based on real microscopic image data, we investigate the capacity of microrobots to collect target cells from within a dissociated tissue culture. Our results indicate the feasibility of using OETdMs to autonomously carry out micromanipulation tasks within complex, unstructured environments.

## I. INTRODUCTION

Optoelectronic tweezer (OET)-driven microrobots (OETdMs) are a recently introduced micromanipulation technology [1] in which light induced dielectrophoresis (DEP) is used to control sub mm-sized actuators (microrobots). These microrobots can be used to carry out a variety of tasks in microscopic environments, such as collecting and transporting objects. Compared to optical tweezers (OT) [2], direct OET can exert significantly larger forces (10s of pN) at substantially lower light irradiance levels, allowing simultaneous manipulation of larger objects [3]. However, the large electric field gradient and heat generated can damage sensitive objects.

OETdMs offer several advantages over direct OET, particularly in the magnitude of the forces they can exert (typically 100s of pN). Significantly, the mechanical forces exerted by microrobots have been shown to be less damaging to mammalian cells [1].

OET techniques are based on the manipulation of an object through the DEP force experienced by a material in a non-uniform electric field. A typical OET device is formed from a pair of parallel plate electrodes, one of which is transparent and the other coated with a photoconductive layer, enclosing a fluid containing the objects of interest [4]. Illuminating a region of the photoconductive layer reduces the local electrical conductivity, introducing a gradient in the electric field in the liquid medium above the dark and illuminated regions when an external potential is applied. As a result, objects are attracted to (positive DEP) or repelled from (negative DEP) illuminated parts of the substrate, depending on the applied field and the relative polarizability of the object and the fluid, as defined by the Clausius–Mossotti factor [5].

In the case of OETdMs, DEP forces are used to control microrobots formed from SU-8 photoresist. The OET electrodes are typically mounted in an optical microscope, with a spatial light modulator (such as a light projector) coupled into the illumination pathway to create light patterns on the photoconductive layer. The microrobots themselves can be fabricated in a range of shapes and sizes, depending on the target application, and in large numbers at low cost using standard photolithography techniques. However, the potential of OETdMs for high throughput, parallel operation remains unrealized due to a reliance on manual control, whereby a trained operator pilots individual microrobots by updating the image sent to the light projector.

In this article we describe an approach which enables simultaneous, unsupervised, open-loop control of multiple OETdMs. We apply low level vision-based detection to first locate target objects, obstacles and microrobots within the field of view of an optical microscope. Microrobot tasks are allocated by modelling the target harvesting task as a capacitated vehicle routing problem (CVRP), modified to accommodate the constraints introduced by the relatively

* Research supported by UK Research and Innovation (grant code ES/T011866/1), the UK Engineering and Physical Sciences Research Council (grant code EP/K005030/1) and the Natural Sciences and Engineering Research Council of Canada (grant code ALLRP 548593-19).

Christopher Bendkowski, Laurent Mennillo, Tao Xu, Vijay Pawar and Danail Stoyanov are with University College London, London, WC1E 6BT, United Kingdom (e-mail: christopher.bendkowski.18@ucl.ac.uk; l.mennillo@ucl.ac.uk; tao.xu.19@ucl.ac.uk; v.pawar@ucl.ac.uk; danail.stoyanov@ucl.ac.uk).

Mohamed Elsayed, Filip Stojic, Harrison Edwards, Shuailong Zhang, Cindi Morshead and Aaron R. Wheeler are with University of Toronto, Toronto, ON M5S 3H6, Canada (e-mail: mohammed.elsayed@mail.utoronto.ca; filip.stojic@mail.utoronto.ca; harrison.edwards@mail.utoronto.ca; shuailong.zhang@utoronto.ca; cindi.morshead@utoronto.ca; aaron.wheeler@utoronto.ca).

Michael Shaw is with University College London, London, WC1E 6BT, United Kingdom and also National Physical Laboratory, Hampton Road, Teddington, TW11 0LW, UK (e-mail: mike.shaw@ucl.ac.uk).

small field of view (FoV) of the microscope. Microrobots are then driven simultaneously along computed paths using a light projector, spatially registered to the microscope camera, to display a series of corresponding illumination patterns on the OET device. We demonstrate the practical performance of the method by automating a simple collect, transport and release operation with silica microspheres using a lab built OETdM system. We then apply computational simulations to investigate the performance of automated OETdMs in a real-world application, based on isolating therapeutically relevant neural stem cells from a dissociated tissue section. A statistical model of the components found in the tissue suspensions, derived from experimental image data, is used to create a set of synthetic environments to test the ability of microrobots to successfully collect target cells whilst avoiding obstacles. Finally, we discuss the implications of our results for the operation of autonomous OETdMs in complex, unstructured environments more generally.

## II. AUTONOMOUS OET DRIVEN MICROROBOTIC SYSTEM

### A. Microscope and OET device configuration

OET devices were prepared as described in [4]. Two electrodes were created by coating a pair of borosilicate glass microscope slides with indium tin oxide (ITO). The lower slide was further coated with a 1 μm thick layer of amorphous hydrogenated silicon (a-Si:H) and the two slides were joined using adhesive acrylate tape to form a sealed chamber 150 μm deep. An electric field (30 kHz, 30 V PP) was applied across the gap using a function generator (U2761A, Agilent) connected to exposed patches of ITO on each slide. Cogwheel shaped microrobots (see inset to Fig. 1) were fabricated from SU-8 using photolithography. After detachment from the silicon substrate upon which they were fabricated, microrobots were suspended in deionized water and added to the a-Si:H layer before assembly of the OET device.

To control and visualize microrobots within the OET devices, we coupled the light output from a commercial digital light projector (LS800HD, ViewSonic) into an upright light microscope (BX51 WI, Olympus) (Fig. 1). The projector was mounted on an optical breadboard, itself supported on aluminum pillars attached to the same active vibration isolation table as the microscope frame. Light from the projector was inserted into the optical path of the microscope using a dual view camera module fitted with a 90/10 beamsplitter. By removing several of the elements in the projector lens assembly, we were able to form a suitably magnified image of the digital micromirror device (DMD) in a plane conjugate to the native image plane (back focal plane of the tube lens). This intermediate image of the DMD was projected onto the OET device substrate by the tube lens and microscope objective lens (UPLANSAPO 4x, Olympus). To reduce the effects of axial chromatic aberration, a long pass optical filter (FD1R, Thorlabs) was inserted into the beam before the beamsplitter. Brightfield images were recorded using a digital camera (Orca Flash 4.0, Hamamatsu Photonics) with a field of view of 3.3 mm x 3.3 mm. With a neutral density filter in front of the camera we were able to simultaneously record an image of the illumination light pattern from the projector (reflected from the a-Si:H substrate), the microrobots, target objects and debris. The position of the OET device itself was controlled using a motorized (XY) translation stage (H117P1XD, Prior Scientific).

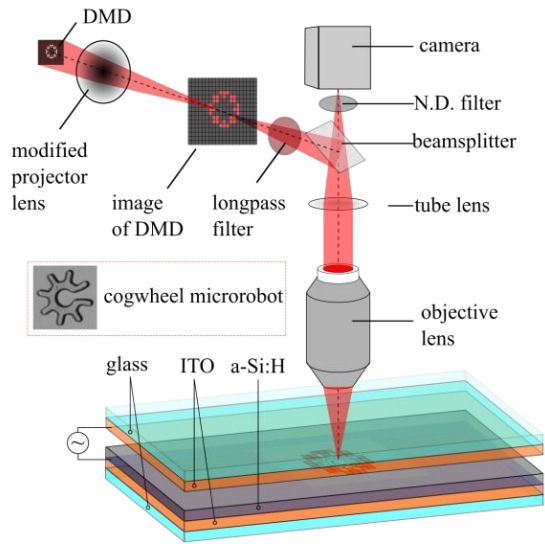

Figure 1. Hardware for visualization and control of OET-driven microrobots. The microrobots and manipulation targets in suspension are inserted into the gap formed between two ITO coated glass slides. The lower slide is also coated with a layer of a-Si:H, the electrical conductivity of which is controlled using light patterns introduced from a coupled digital light projector. When an external potential is applied the resulting non-zero transverse electric field gradient within the gap is used to manipulate the robots. Figure inset shows a brightfield image of a cogwheel shaped microrobot with an outer diameter of 200 μm.

Image sprites (light patterns) to trap, rotate and translate microrobots were generated using the python multimedia library pyglet [6]. Brightfield images from the camera were captured using the MicroManager [7] driver with its python API. The motorized translation stage was controlled using serial commands.

### B. Camera and projector calibration

In order to project a light pattern at the desired location in the camera image, which is used for object detection, we projected an alternating pattern of rows and columns of dots using the DMD projector while capturing images with the camera. Detected locations of the projected dots in image coordinates were then correlated to their associated DMD coordinates and interpolated to give a full mapping from the camera coordinates to the DMD coordinates.

### C. Microrobot and target detection

Precise determination of the position and orientation of microrobots and the position of targets (silica microspheres) within the field of view of the camera was accomplished using two separate detection algorithms based on traditional image processing techniques implemented in OpenCV [8] and scikit-image [9] libraries. Additional detected objects were labelled as debris / obstacles, which microrobots were required to avoid during navigation.

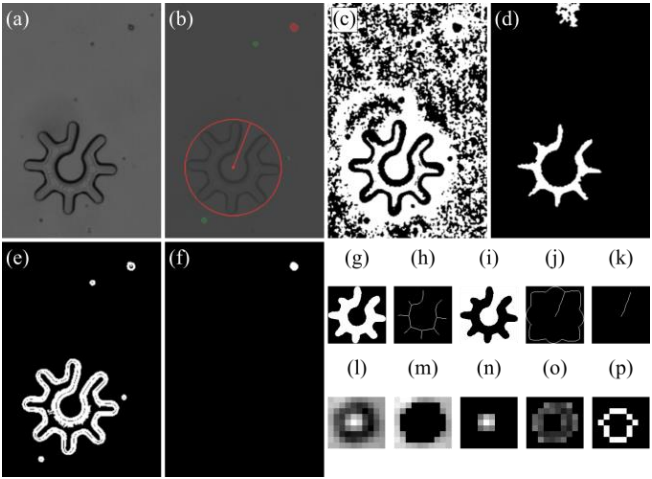

Figure 2. Microrobot and microsphere detection. (a) Original brightfield image. (b) Final result of the detection pipeline showing the detected microrobot (red), microsphere (red) and debris (green). (c) Microrobot detection in intensity mode after binarization. (d) Microrobot candidate in intensity mode before morphological dilation. (e) Edge thresholding for microsphere detection. (f) Detected microsphere candidate. (g) Microrobot candidate in intensity mode, (h) its skeleton, (i) inverse image, (j) skeleton of its inverse image and (k) largest skeleton branch after contour removal. (l) Grayscale image of microsphere candidate, (m) outer area, (n) inner area, (o) ring and (p) skeleton of ring mask.

*1) Microrobot detection*

The pipeline for the detection of microrobots is composed of three stages, with an initial stage operating under two distinct modes that can be used jointly to improve the rate of detection. In the first detection mode, a threshold on the image intensities is used to detect the brighter, semi-enclosed chamber at the center of each microrobot, which allows for the detection of robots in contact with other objects, while the second mode relies on edge detection to define their outer silhouette, which is more robust for the detection of isolated robots that have collected debris or microspheres. In the first detection mode, adaptive histogram equalization [10] and Gaussian filtering were applied before binarization with hysteresis thresholding (Fig. 2(c)). Morphological opening and closing operations were then performed to ensure the inner part of each robot was unconnected to any other structure. Very small and very large connected components were removed (Fig. 2(d)), and all remaining components were individually filled and slightly morphologically dilated. In the second detection mode, adaptive histogram equalization was followed by edge detection using Canny's algorithm [11]. A morphological closing operation was then performed on the detected edges, before filling of each connected component and removal of those touching the image borders. Finally, all remaining connected components were classified as microrobot candidates.

To identify real microrobots among these candidates, a series of shape and intensity constraints were applied. Firstly, an upper threshold ($T_{ar}$) was set for the ratio of the major axis length $a_M$ over minor axis length $a_m$ of the ellipse with the same normalized second central moments as the component ($a_M/a_m < T_{ar}$). Secondly, the solidity $s$ of the real microrobots was required to lie between two thresholds $T_{sl} < s < T_{sh}$. Thirdly the number of branches on the skeleton of the component was required to be greater than or equal to five (Fig. 2(h)). Although microrobots are fabricated with six protuberances setting this lower threshold improved detection reliability. Finally, an intensity constraint was applied to ensure that the center of the candidate microrobot was empty. From empirical testing best results were obtained using threshold values of $T_{ar} = 1.30$, $T_{sl} = 0.45$ and $T_{sh} = 0.75$. For every candidate satisfying these constraints, its center and radius were estimated by computing the minimum enclosing circle of its connected component (Fig. 2(g)). The orientation was computed from the slope of the line fitted on the pixel coordinates of the longest branch of its skeletonized inverse image (Fig. 2(i)-(j)), after removal of the outer contour (Fig. 2(k)).

*2) Microsphere and debris detection*

For experimental demonstration, we carried out a simple collect, transport and release operation using OETdM to swap the positions of 10 μm diameter silica microspheres. In the brightfield images, these microspheres appeared as a dark circular ring surrounding a bright central core (Fig. 2(l)). Microspheres were detected using four priors: size, roundedness, contour intensity, and ring shape. After Gaussian smoothing of the original image, gradients were detected using a Scharr filter [12] and enhanced with adaptive histogram equalization. Two different hysteresis thresholds were then applied to retain the microsphere structures at the higher threshold and all other structures at the lower threshold. Following morphological closing and filling of the connected components in the higher threshold binary image, a morphological opening was performed, very large and very small components were removed, and the remaining structures were skeletonized. Components with larger 8-connected skeletons, not compatible with the rounded shape prior of the spheres, were removed, leaving a set of microsphere candidates (Fig. 2(f)). These candidates were then used as masks and individually processed to identify real microspheres as those conforming to a ring shape. For each mask, slight Gaussian smoothing of the corresponding original grayscale image (Fig. 2(l)) was performed, before an inverse binarization step using Otsu's thresholding algorithm [13]. The resulting structure was then filled and inverted again to obtain a mask of the outer area surrounding the bead (Fig. 2(m)), while the intensities of its complement were also binarized to separate the brighter microsphere center area (Fig. 2(n)) from the darker ring (Fig. 2(o)), producing two other masks. Three constraints were then applied to the skeleton of the ring mask (Fig. 2(p)), which should have a unique 8-connected component, only two contours (inner and outer) and no branches. A final constraint was then applied to ensure the mean intensity value of the ring area was lower than the mean intensity values of the outer and center areas. True microspheres (targets) were then identified as candidates satisfying these constraints. Finally, debris labels were obtained by morphological closing of the lower threshold binary image, binary filling of the connected components, removal of the very small and very large ones, and subtraction of the segmented microsphere objects.

*D. Multi-robot path planning*

The task of harvesting objects (cells or microspheres) in a microscopic environment shares similarities with other gathering tasks but also presents new challenges. Considering a known environment – where the locations of targets, obstacles and robots have been determined using the methods described previously (section II C) – the harvesting task can be modelled as a capacitated vehicle routing problem (CVRP). The CVRP is a form of the multiple traveling salesman problem wherein the vehicles each have a capacity constraint $Q$. For the harvesting task, the microrobots originate from one or more depots, collect a maximum of eight targets (their capacity for the 10 μm diameter microspheres used in the following experimental demonstrations) and return to a depot where the targets are released. We seek to minimize the total distance travelled by all microrobots, using Euclidean distance as a heuristic. In contrast to traditional CVRP, where vehicles may travel anywhere in the environment independently of each other, in OETdM the working area ($A_w$) over which microrobots can be controlled is limited to the region within the field of view (FoV) of the microscope where light can be projected. To model this constraint within the CVRP framework, we treated $m$ microrobots as a single vehicle $s$, with $Q_s = \sum_{i=0}^{m} Q_i$. Target objects were grouped into clusters with size $n \leq Q_s$, such that each group lay within an area $A_{cluster} \leq A_w$. K-means clustering was used to find clusters that satisfied $A_{cluster} \leq A_w$. These clusters were then split recursively until $n \leq Q_s$. The formulated CVRP was then solved using the optimizer from the Google OR-Tools library [14]. The solution provided an ordered list of target clusters and depots accessible to all microrobots. The individual microrobots were then assigned to the closest target in the cluster based on Euclidean distance. Where $n < m$, goals for the microrobots without a target were generated using Poisson-disk sampling, ensuring that their goals did not collide and remained within the working area.

The CVRP solution sets the start and goal locations for the journeys undertaken by each microrobot. To plan the paths to complete these journeys requires a Multi-Agent Path Finding (MAPF) algorithm. These algorithms can be described in terms of completeness – the ability of the algorithm to find a solution if one exists, and optimality – the ability of the algorithm to optimize the solution according to some objective function (e.g. makespan). Optimal and complete algorithms are broadly categorized into four approaches: A*-based, Constraint Programming (CP), Conflict Based Search (CBS) and Increasing Cost Tree Search (ICTS). It is generally accepted that A*-based and CP approaches are more suited to small graphs, whereas CBS and ICTS are better for larger graphs [15]. In the case of OETdMs the harvesting task generates large sparse graphs, therefore it was decided that CP and A*-based approaches were not suitable for our application.

In addition to graph size, another constraint is that agents in the MAPF algorithm must be able to occupy more than one vertex in the graph because microrobots are many times larger than the targets and surrounding obstacles. Li et al. proposed CBS and CP based approaches [16], however their experimental results were limited to a small (194 x 194) grid with 51.3% occupancy and although the speed of their algorithm at this scale improved on previous approaches, our graphs are an order of magnitude larger. Atzmon et al. demonstrated a continuous time CBS approach [17], which also accommodates agents of different sizes, but is restricted to agents that occupy at most one vertex. For path planning in OETdM, we adopted the Any-Angle Safe Interval Path Planning (AA-SIPP(m)) approach [18]. Adapted from the single-agent path planning method, Safe Interval Path Planning (SIPP) [19], AA-SIPP(m) is able to avoid dynamic obstacles and uses prioritized planning. In AA-SIPP(m), the paths of agents are planned in the order of their assigned priority, with lower priority agents required to navigate to avoid higher priority ones. One downside is that AA-SIPP(m) is only complete under a restricted set of circumstances and does not guarantee optimality. The benefit of such a decoupled approach, however, is its computational efficiency and we were able to solve problems on our large graphs in a reasonable amount of time (<1 s for simple routing problems and up to 300 s when solving scenarios with large distances between microrobots and targets). To accommodate the limited working area ($A_w$) in our OETdM setup, we altered the lower-level path finding within AA-SIPP(m) to avoid paths which would violate this constraint. The performance of the algorithm in the case of uniform (identical) robots was improved by only considering vertices occupied by the perimeter of the robots when detecting collisions.

### E. Experimental results

To experimentally demonstrate the autonomous capabilities of our OETdM system, we applied it to a simplified reordering task involving silica microspheres. In these experiments, two or three microrobots were paired with corresponding microspheres and assigned the task of swapping microsphere positions on the surface of the lower OET electrode. This process was split into three stages, collect, transport and release, with every microrobot required to finish each stage before any of the microrobots moved on to the next. Fig. 3 shows frames from a time lapse video of three microrobots successfully completing the task (full execution of the task is shown in the supplementary video).

As our system is based on open-loop control of the microrobots, it is unable to accommodate any significant motion of the targets between the target detection and collection stages. In many applications, such as collecting and sorting biological cells, interaction forces tend to stabilize the position of the target objects and the substate [20]. However, in the case of microspheres, we found that the targets tended to drift across the substrate and were also attracted to microrobots during approach. To mitigate this issue, we stabilized the microspheres using OET by projecting an annulus light pattern around them during the collect and release of the task.

While the maximum translation speed of the microrobots in our setup was 1100 μms$^{-1}$, we used a reduced translation speed of 65 μms$^{-1}$ to improve the reliability of the microsphere swapping task. At higher speeds we occasionally found microspheres became trapped under microrobots during the transport stage. We investigated the failure modes to understand why a reduced speed was necessary. Firstly, our finite element simulations (data not shown) indicate there is a vertical component to the DEP force acting on the microrobots which we believe raise them slightly above surface of the a-Si:H layer. This creates a small gap under which microspheres can become stuck. Secondly, each microrobot performs slightly differently, likely due to inconsistencies in manufacture and handling, with some able to achieve higher velocities than others. Thirdly, we measured maximum velocity, by fixing the microrobots in position using OET and translating the OET device underneath them using the motorized stage. Conversely, in the microsphere swapping experiments, the stage and OET device remained fixed and the light pattern was moved across the working area. This likely reduced the achievable velocity as the intensity and contrast of the projected image can vary across the working area, resulting in non-uniform DEP forces and reduced control.

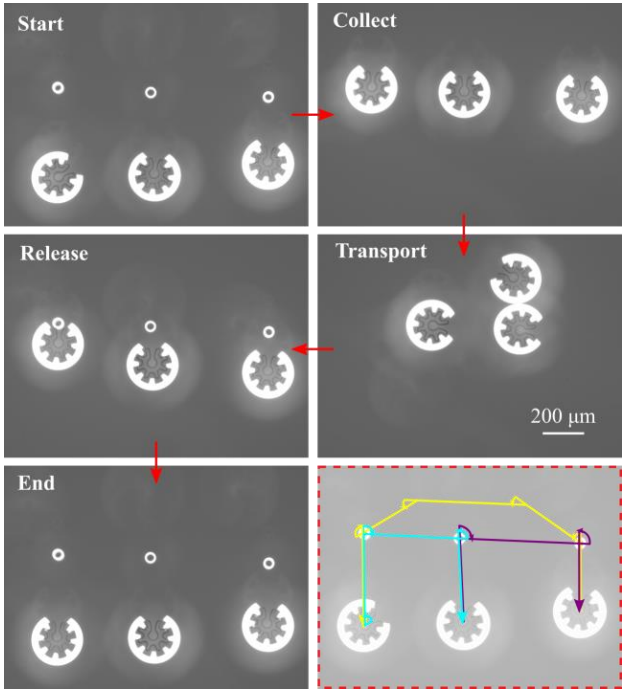

Figure 3. Simultaneous automated collection, transport and release of three silica microspheres using OETdM. Lower right panel shows the complete paths followed by the three microrobots.

## III. SIMULATION OF CELL HARVESTING USING OETdM

To assess the performance of our method for autonomous cell harvesting, we developed an OETdM simulation based on microscopic images of dissociated murine brain tissue. A small sub-region of the brain, known as the subventricular zone, contains therapeutically promising stem cells and their progeny, collectively known as neural precursor cells (NPCs) [21]. Isolating the small number of NPCs, from other cells and tissue components is a task potentially well suited to OETdMs, particularly as existing methods (such as fluorescence activated cell sorting) are inefficient and lack specificity.

### A. Simulation of dissociated tissue environment

A generative model, simulating the shapes and distributions of the observed microscopic structures (cells and other tissue components), was generated from a series of brightfield images of dissociated SVZ resections. The model generates synthetic images of target cells (NPCs) and debris structures at different concentrations to allow for the evaluation of the path planning framework in controlled and realistic conditions. The first stage involves image segmentation and ground truth labelling, followed by a second stage which models and clusters the cell and debris label shapes. The third stage models the distribution of these shape clusters and the overall label density per image. Finally, the last stage generates synthetic images from the computed shape and distribution models.

#### 1) Image segmentation and ground truth labelling

Image segmentation was performed using a similar method to that described previously for microsphere detection (see section II C), up to the point at which the ring shape prior is evaluated to identify real microspheres among candidates, which were considered here as the segmented cells labels. However, two changes were made to the detection pipeline. Instead of smoothing the original image, exposure correction was computed by dynamically adjusting the gamma value for each image in order to produce corrected images with the same mean intensity level, and the morphological opening operation was removed. Debris labels were also computed following the same procedure as that described earlier (see section II C), with an additional morphological opening operation after the closing operation. Alongside automatic segmentation, ground truth cell labels (indicating real cells coordinates) were also computed from annotations provided by experts. The F1 score of the automatic cell segmentation obtained on a sample of 156 images at a concentration of 400 cells/µL was 0.59. An example ground truth labelled image is shown in Fig. 4(b).

#### 2) Modelling and clustering label shapes

The second stage of the generative model performs joint modelling and clustering of the ground truth label shapes. Several shape characteristics were evaluated for each label, or region, in all images: the area, major and minor axis lengths of the ellipse that has the same normalized second central moments as the region, solidity, thickness, fiber length and number of skeleton branches. As the ground truth cell labels were known beforehand, clustering was only performed on debris labels, while shape models were computed for both label types. Considering the shape characteristics $s = 7$, the shapes of a cell region $X_{cell} \in \mathbb{R}^s$ were modelled by a multivariate normal distribution

$$X_{cell} \sim \mathcal{N}(\mu, \Sigma). \quad (1)$$

Debris regions $X_{debris} \in \mathbb{R}^s$ were modelled using a Gaussian Mixture Model (GMM)

$$p(X_{debris}) = \phi_i \sum_{i=1}^{K} \mathcal{N}(X_{debris}|\mu_i, \Sigma_i), \quad (2)$$

with $K=6$ the number of GMM components, or debris shape classes, and $\phi_i$ equal prior mixture weights. These two models were fitted with Maximum Likelihood Estimation (MLE) using the scikit-learn library [22]. Example clustered labelled images are shown Fig. 3(c) and (f).

#### 3) Modelling the distribution of clustered labels

Similar to shape models, the distribution $X_{dist}^c \in \mathbb{R}^{C_s}$ of each shape class $C_s = 7$ (one cell class plus six debris classes) per image at concentration $c$ were modelled by a multivariate normal distribution fitted with MLE,

$$X_{dist}^c \sim \mathcal{N}(\mu, \Sigma). \quad (3)$$

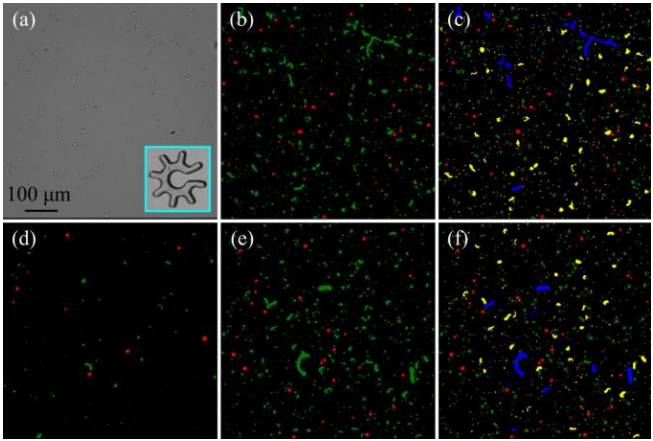

Figure 4. Synthetic image generation. (a) Typical brightfield image of a cell suspension, inset shows an individual cogwheel microrobot. (b) The same image labelled to show target (neural) cells in red and debris in green. (c) Clustered image with target cells in red and debris belonging to different shapes classes represented in different colors. (d) Example synthetic image corresponding to 100 cells/μL. (e) Example synthetic image corresponding to 400 cells/μL. (f) The same image as in (e) with debris colored according to shape class.

Our dataset consisted of seven image sets acquired at different cell concentrations and seven distribution models were fitted for each image set, at 1, 10, 20, 50, 100, 200 and 400 cells/μL. In addition to these distribution models, a cubic function $f: \mathcal{C} \mapsto \mathcal{L}$ was fitted on the number of labels per image $\mathcal{L}$ for the whole dataset to evaluate the model for concentrations in the continuous interval $\mathcal{C} = [1, 400]$.

*4) Synthetic image generation*

Finally, synthetic images were generated using the computed shape and distribution models. After setting a desired output concentration $c$, the first step of this stage consisted of selecting the distribution model with the nearest concentration value as a reference. This model was then used to estimate the number of regions to generate for each shape class. Image generation operated in two distinct modes: a uniform density mode and a variable density mode. In uniform density mode, the image was generated in a single step, with the number of regions for each shape class equal to the component means of the selected distribution model. To accommodate different concentrations, the number of regions was multiplied by a density factor computed from the value of $f$ at the desired concentration over the sum of all means of the distribution model. Finally, another scaling factor was applied to manually adjust the final label density. This resulted in final synthetic images with a stable number of regions for each shape class at the set concentration. In variable density mode, the output image was divided into a grid for which the distribution of each tile was directly sampled from the selected distribution model. This mode tended to generate more realistic results, at the expense of a less controllable label density per image. In the case of output images of size different to those of the training dataset, the distribution was also multiplied by an area factor computed to maintain the same label density regardless of image size. The computed image distribution, or tile distributions in variable density mode, was then used to generate regions corresponding to their respective shape class. Each region was generated from a sample of its specific shape class component to retrieve a set of sampled shape values. A random spline was generated from the sampled fiber length and solidity values, while the sampled number of branches was used to generate other splines starting from random points of the first spline. Ellipses on random points of the splines were then drawn, with major and minor axis lengths computed from the sampled thickness value. Finally, the whole region was dilated until its area matched the sampled area value. Once all regions had been generated, they were finally drawn without overlap on the final image at random coordinates, following a uniform spatial distribution. Examples of generated images at different cell concentrations are shown Fig. 4(d)-(f).

*B. Simulation results*

To demonstrate the capabilities of our task assignment and path-planning approach, we simulated harvesting tasks in environments with varied densities. The simulated environments consisted of a 1 cm x 1 cm device, populated with target cells and debris. In each simulation, microrobots started at the center of the field within an empty 1 mm x 1 mm square. Using the CVRP solver, microrobots were assigned target cells to collect and return to the starting area, with journeys planned using AA-SIPP(m). Fig. 5 shows the rate of successful journeys for up to five microrobots compared with the amount of free space in the environment, where free space is defined as the space in the environment that a microrobot can occupy without colliding with obstacles. We found journeys were not completed either due to invalid starting/goal positions or path-planning failures. Invalid start/goal positions occur when the targets are not reachable without microrobots colliding with another obstacle, and account for 60-100% of all failures with two or more microrobots, and 25-47% of all failures in single microrobot experiments. The disparity between these figures arises because all microrobots must be assigned goals in the same working area and the probability of having sufficient free space diminishes with each additional microrobot. Path planning failures also increase with the number of microrobots in the experiment, as the difficulty of finding non-conflicting, passing paths and keeping all paths within the working area increases.

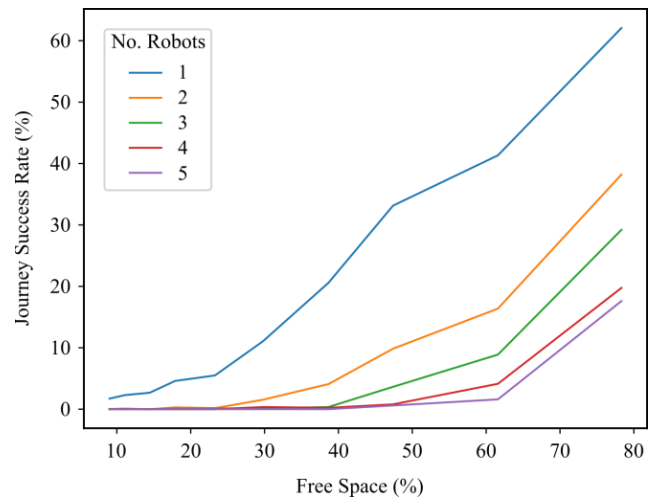

Figure 5. Simulated results showing journey success rate for different numbers of microrobots versus free space percentage for OETdM in a dissociated tissue culture. Free space is defined as the space in the environment that a microrobot can occupy without collision.

For real world context, our object detection analysis (section III A) suggests a dissociated neural tissue suspension with a concentration of 1 cell/µL corresponds to 1.1% free space on the device substrate. The results in Fig. 5 indicate that a significantly lower object density is required for microrobots to navigate effectively. This could be achieved by further dilution of the cell suspension before loading. In practice, we find microrobots can be navigated effectively at somewhat higher suspension concentrations as the light pattern used to trap them tends to repel some debris. In addition, the microrobots are still able move with small amounts of debris adhered to their outer edge. An alternate strategy would be to intentionally use either direct OET, or a subset of microrobots to clear paths through the debris, which is an area of active investigation.

## IV. CONCLUSION

Optoelectronic tweezer-driven microrobots (OETdMs) are a useful and versatile tool for manipulating micro and nano-sized objects. Our work outlines for the first time an approach for automation of OETdM control, from the detection of microrobots and targets in brightfield microscopic images, task allocation path planning to control of microrobots in an OET device by updating calibrated light patterns projected onto the photoconductive electrode. These technologies were demonstrated experimentally using OETdMs to automate a simple microsphere swapping task. To assess the feasibility of using this method for cell harvesting, we developed a generative model based on experimental images of dissociated murine neural tissues. The results suggest that our task allocation and path planning approach is effective in lower density samples, however, alternative strategies will likely be required for automation of OETdMs in environments with a higher concentration of targets and obstacles.

Our work, while a step towards automation, also reveals the need for further work to completely automate OETdM control for complex tasks such as cell harvesting. The focus of our future work in this area will include the developments of new models to simulate the interaction between forces in this complex fluid environment, closed-loop control of microrobots using visual servoing, new robot designs and differential task assignment using a mixed microrobot population. Finally, deep neural networks offer significant potential for more reliable detection of target cells in complex, multi-component environments; reinforcement learning approaches are particularly promising for robot navigation and task allocation.